%% file: main.tex
\documentclass[11pt, a4paper, logo, copyright, nonumbering]{deepseek}
\usepackage[authoryear, sort&compress, round]{natbib}
\usepackage{dblfloatfix}
\usepackage{ulem}
\usepackage{caption}
\usepackage{subcaption}
\usepackage{makecell}
\usepackage[utf8]{inputenc}
\usepackage{dramatist}
\usepackage{xspace}
\usepackage{pifont}
\usepackage{multirow}
\usepackage{tcolorbox}
\usepackage{tabularx}
\usepackage{xltabular}
\usepackage{longtable}
\usepackage{hyperref}
\AtBeginDocument{

}
\hypersetup{
    colorlinks=true,
    linkcolor=blue,
    citecolor=blue,
    urlcolor=blue,
    filecolor=blue
}
\usepackage{xcolor}
\usepackage{array}
\usepackage{amsfonts}
\usepackage{amsmath}
\usepackage{amssymb}
\usepackage{lineno}
\usepackage{bm}
\usepackage{booktabs}
\usepackage{ragged2e}

\usepackage[bottom]{footmisc}

\usepackage{CJKutf8}
\usepackage{setspace}

\usepackage{dsfont}

\makeatletter
\def\@BTrule[#1]{%
  \ifx\longtable\undefined
    \let\@BTswitch\@BTnormal
  \else\ifx\hline\LT@hline
    \nobreak
    \let\@BTswitch\@BLTrule
  \else
     \let\@BTswitch\@BTnormal
  \fi\fi
  \global\@thisrulewidth=#1\relax
  \ifnum\@thisruleclass=\tw@\vskip\@aboverulesep\else
  \ifnum\@lastruleclass=\z@\vskip\@aboverulesep\else
  \ifnum\@lastruleclass=\@ne\vskip\doublerulesep\fi\fi\fi
  \@BTswitch}
\makeatother

\addto\extrasenglish{
}

 {\begin{list}{}%
         {\setlength{\leftmargin}{#1}}%
         \item[]%
 }
 {\end{list}}
 
\reportnumber{001} 

\renewcommand{\today}{}

\title{
\centering Conditional Memory via Scalable Lookup: \\ A New Axis of Sparsity for Large Language Models
}

\author[]{
Xin Cheng$^{1,2*}$, Rui Tian$^{2*}$, Wangding Zeng$^{2}$, Damai Dai$^{2}$, Qinyu Chen$^{2}$, 

\vspace{-0.1in}
Bingxuan Wang$^{2}$, Zhenda Xie$^{2}$, Kezhao Huang$^{2}$, Xingkai Yu$^{2}$

\vspace{-0.1in}
Chengqi Deng$^{2}$, Shangyan Zhou$^{2}$, Chenggang Zhao$^{2}$, Zhewen Hao$^{2}$

\vspace{-0.1in}
Yukun Li$^{2}$, Han Zhang$^{2}$, Zhengyan Zhang$^{2}$, Yixuan Wei$^{2}$, M.Y Xu$^{2}$

\vspace{-0.1in}
Huishuai Zhang$^{1}$, Dongyan Zhao$^{1}$, Wenfeng Liang$^{2}$ \\
\vspace{-0.1in}
\small
$^1$Peking University  \quad $^2$DeepSeek-AI \\
\small
\{zhanghuishuai, zhaody\}@pku.edu.cn \\
\small 
\{chengxin, tianr22, zengwangding, damai.dai\}@deepseek.com

\vspace{-0.2in}
}

\input{commands.tex}

\begin{abstract}
While Mixture-of-Experts (MoE) scales capacity via conditional computation, Transformers lack a native primitive for knowledge lookup, forcing them to inefficiently simulate retrieval through computation. To address this, we introduce conditional memory as a complementary sparsity axis, instantiated via \textbf{Engram}, a module that modernizes classic $N$-gram embedding for $\mathcal{O}(1)$ lookup.
By formulating the \textit{Sparsity Allocation} problem, we uncover a U-shaped scaling law that optimizes the trade-off between neural computation (MoE) and static memory (Engram).
Guided by this law, we scale Engram to 27B parameters, achieving superior performance over a strictly iso-parameter and iso-FLOPs MoE baseline. 
Most notably, while the memory module is expected to aid knowledge retrieval (e.g., MMLU $+3.4$; CMMLU $+4.0$), we observe even larger gains in general reasoning (e.g., BBH $+5.0$; ARC-Challenge $+3.7$) and code/math domains~(HumanEval $+3.0$; MATH $+2.4$).
Mechanistic analyses reveal that Engram relieves the backbone's early layers from static reconstruction, effectively deepening the network for complex reasoning. Furthermore, by delegating local dependencies to lookups, it frees up attention capacity for global context, substantially boosting long-context retrieval (e.g., Multi-Query NIAH: $84.2 \to 97.0$).
Finally, Engram establishes infrastructure-aware efficiency: its deterministic addressing enables runtime prefetching from host memory, incurring negligible overhead.
We envision conditional memory as an indispensable modeling primitive for next-generation sparse models. Code available at: \url{https://github.com/deepseek-ai/Engram}.
\end{abstract}

\begin{document}
\begin{CJK*}{UTF8}{gbsn}

\maketitle
\renewcommand{\thefootnote}{*}
\footnotetext{Equal contribution.}
\renewcommand{\thefootnote}{\arabic{footnote}}

\input{sections/intro}
\input{sections/arch}
\input{sections/scaling_law_exp}
\input{sections/large_scale_exp}

\input{sections/long_context_exp}
\input{sections/analysis}
\input{sections/related_work}

\section{Conclusion}

In this work, we introduce \textbf{conditional memory} as a complementary sparsity axis to the prevailing conditional computation paradigm (MoE), aiming to resolve the inefficiency of simulating knowledge retrieval through dynamic computation. We instantiate this concept via Engram, a module that modernizes classic $N$-gram embeddings to enable scalable, constant-time $O(1)$ lookups for static patterns.

By formulating the \textit{Sparsity Allocation} problem, we uncover a U-shaped scaling law, demonstrating that a hybrid allocation of sparse capacity between MoE experts and Engram memory strictly outperforms pure MoE baselines. Guided by this law, we scale Engram to 27B parameters, achieving superior performance across diverse domains. Notably, while the memory module intuitively aids knowledge retrieval, we observe even larger gains in general reasoning, code, and mathematics.

Our mechanistic analysis reveals that Engram effectively ``deepen'' the network by relieving early layers from static reconstruction tasks, thereby freeing up attention capacity to focus on global context and complex reasoning. This architectural shift translates into substantial improvements in long-context capabilities, as evidenced by performance gains in LongPPL and RULER. Finally, Engram advocates for infrastructure-aware efficiency as a first-class design principle. Its deterministic addressing allows for the decoupling of storage and compute, enabling the offloading of massive parameter tables to host memory with negligible inference overhead. We envision conditional memory functions as an indispensable modeling primitive for next-generation sparse models.

\bibliography{ref}

\newpage
\appendix

\section*{Appendices}
\section{Detailed Model Architecture and Hyper Parameters}
\label{appendix:detailed_model_arch}
\input{tabs/appendix_detailed_arch}

\clearpage
\section{Full Benchmark Curves}
\label{appendix:benchmark_curve}
\begin{figure}[htbp]
    \centering
    \includegraphics[height=0.74\textheight, keepaspectratio]{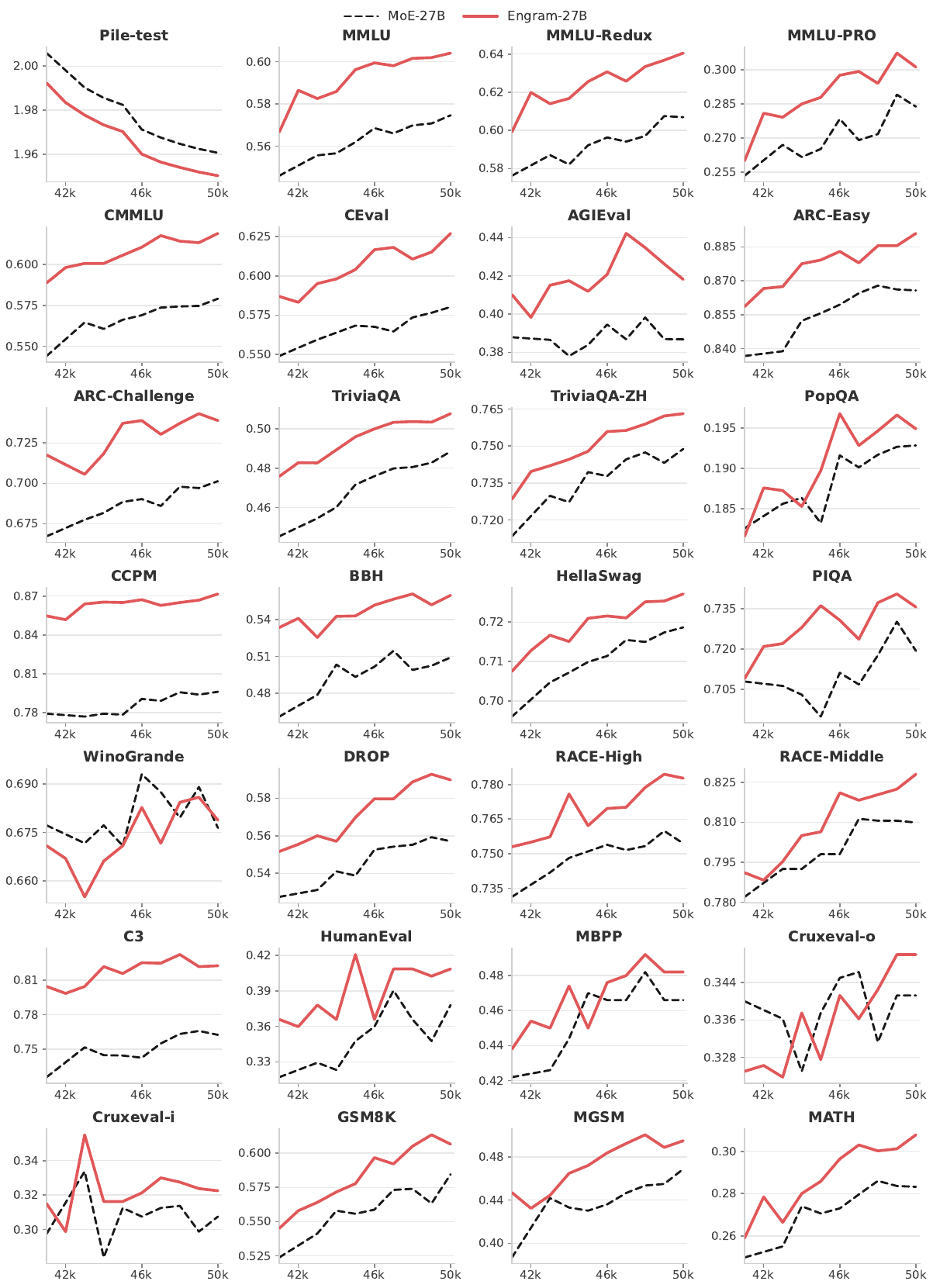}
    \caption{Last 10k pre-training benchmark curve.}
    \label{fig:benchmark_curve}
\end{figure}
\clearpage
\section{Case Study of Tokenizer Compression}
\label{appendix:token_compression}

\newcommand{\spacebox}{\textvisiblespace}
\begin{table}[h!]
    \centering
    \label{tab:token_collisions}
    \small
    \begin{tabularx}{\textwidth}{c c c X}
        \toprule
        \textbf{Rank} &  \textbf{\makecell{Merge\\Count}} & \textbf{\makecell{Normalized\\Token}} & \multicolumn{1}{c}{\textbf{Original Tokens}} \\
        \midrule
        
        1 & 163 & \texttt{'\spacebox'} & \texttt{'\textbackslash t'}, \texttt{'\textbackslash n'}, \texttt{'\textbackslash r'}, \texttt{'\spacebox'}, \texttt{'\spacebox\spacebox'}, \texttt{'\textbackslash n\textbackslash n'}, \texttt{'\spacebox\spacebox\spacebox'}, \texttt{'\spacebox\textbackslash n'}, ... \\
        \addlinespace[0.5ex]
        
        2 & 54 & \texttt{'a'}  & \texttt{'A'}, \texttt{'a'}, \texttt{'\spacebox a'}, \texttt{'\spacebox A'}, \texttt{'á'}, \texttt{'ä'}, \texttt{'ã'}, \texttt{'ą'}, \texttt{'\spacebox à'}, \texttt{'\spacebox å'}, \texttt{'â'}, ... \\
        \addlinespace[0.5ex]
        
        3 & 40 & \texttt{'o'}  & \texttt{'O'}, \texttt{'o'}, \texttt{'\spacebox o'}, \texttt{'\spacebox O'}, \texttt{'ó'}, \texttt{'ö'}, \texttt{'ô'}, \texttt{'õ'}, \texttt{'ő'}, \texttt{'ò'}, ... \\
        \addlinespace[0.5ex]
        
        4 & 35 & \texttt{'e'}  & \texttt{'E'}, \texttt{'e'}, \texttt{'\spacebox e'}, \texttt{'\spacebox E'}, \texttt{'é'}, \texttt{'è'}, \texttt{'\spacebox é'}, \texttt{'ę'}, \texttt{'ě'}, \texttt{'ê'}, ... \\
        \addlinespace[0.5ex]
        
        5 & 30 & \texttt{'i'}  & \texttt{'I'}, \texttt{'i'}, \texttt{'\spacebox I'}, \texttt{'\spacebox i'}, \texttt{'í'}, \texttt{'ì'}, \texttt{'î'}, \texttt{'ī'}, \texttt{'ï'}, ... \\ 
        \addlinespace[0.5ex]
        \bottomrule
    \end{tabularx}
    \caption{The table illustrates Top-5 merged tokens by \textit{Tokenizer Compression} and the overall compression ratio is 23.43\% for our 128k tokenizer.}
\end{table}

\clearpage

\end{CJK*}
\end{document}

%% file: commands.tex
\renewcommand{\phi}{\varphi}

\renewcommand{\epsilon}{\varepsilon}
\renewcommand{\imath}{\mathrm{i}}

\newlength{\restsubwidth}
\newlength{\restsubheight}
\newlength{\restsubmoreheight}
\newcommand{\rest}[2]{%
        \settowidth{\restsubwidth}{\ensuremath{#2}}
        \settoheight{\restsubheight}{\ensuremath{{}_{#2}}}
        \ensuremath{{#1\hskip 0.5pt}_{\vrule\kern2pt\parbox[b][%
        4pt][b]{\the\restsubwidth}{%
                        \ensuremath{{}_{#2}}}}}
        }

%% file: sections/intro.tex
\section{Introduction}
\label{sec:intro}
Sparsity is a recurring design principle for intelligent systems, spanning from biological neural circuits~\citep{olshausen1997sparse,lennie2003cost} to modern Large Language Models (LLMs). Currently, this principle is primarily realized through Mixture-of-Experts (MoE)~\citep{shazeer2017outrageously,dai2024deepseekmoe}, which scales capacity via conditional computation. 
Owing to its ability to drastically increase model size without proportional increases in compute, MoE has become the de facto standard for frontier models~\citep{guo2025deepseek,comanici2025gemini,kimi-k2}.

Despite the success of this conditional computation paradigm, the intrinsic heterogeneity of linguistic signals suggests significant room for \textit{structural optimization}. Specifically, language modeling entails two qualitatively different sub-tasks: compositional reasoning and knowledge retrieval. While the former demands deep, dynamic computation, a substantial portion of text—such as named entities and formulaic patterns—is local, static, and highly stereotyped~\citep{erman2000idiom,constant2017survey}. The effectiveness of classical $N$-gram models~\citep{liu2024infini,nguyen2024understanding,brants-etal-2007-large} in capturing such local dependencies implies that these regularities are naturally represented as computationally inexpensive lookups. Since standard Transformers~\citep{vaswani2017attention} lack a native knowledge lookup primitive, current LLMs are forced to \textbf{simulate retrieval through computation}. For instance, resolving a common multi-token entity requires consuming multiple early layers of attention and feed-forward networks~\citep{ghandeharioun2024patchscopes,DBLP:conf/coling/JinYHZWH0MMDYDZ25} (see \autoref{tab:patchscope}). This process essentially amounts to an expensive runtime reconstruction of a static lookup table, wasting valuable sequential depth on trivial operations that could otherwise be allocated to higher-level reasoning.

To align model architecture with this linguistic duality, we advocate for a complementary axis of sparsity: conditional memory. Whereas conditional computation sparsely activates parameters to process dynamic logic~\citep{shazeer2017outrageously,bengio2013estimatingpropagatinggradientsstochastic}, conditional memory relies on sparse lookup operations to retrieve static embeddings for fixed knowledge. As a preliminary exploration of this paradigm, we revisit $N$-gram embeddings~\citep{bojanowski2017enriching} as a canonical instantiation: local context serves as a key to index a massive embedding table via constant-time $\mathcal{O}(1)$ lookups~\citep{tito2017hash,huang2025over,pagnoni2025byte,yu2025scaling}. Our investigation reveals that, perhaps surprisingly, this static retrieval mechanism can serve as an ideal complement to modern MoE architecture—but only if it is properly designed. In this paper, we propose \textbf{Engram}, a conditional memory module grounded in the classic $N$-gram structure but equipped with modern adaptations such as tokenizer compression, multi-head hashing, contextualized gating, and multi-branch integration~(detailed in \autoref{sec:architecture}).

To quantify the synergy between these two primitives, we formulate the \textit{Sparsity Allocation} problem: given a fixed total parameter budget, how should capacity be distributed between MoE experts and Engram memory? Our experiments uncover a distinct U-shaped scaling law, revealing that even simple lookup mechanisms, when treated as a first-class modeling primitive, act as essential complements to neural computation. 
Guided by this allocation law, we scale Engram to a 27B-parameter model. Compared to a strictly iso-parameter and iso-FLOPs MoE-27B baseline, Engram-27B achieves superior efficiency across diverse domains. Crucially, the gains are not limited to knowledge-intensive tasks (e.g., MMLU: $+3.4$; CMMLU: $+4.0$; MMLU-Pro: $+1.8$), where memory capacity is intuitively beneficial; we observe even more significant improvements in general reasoning (e.g., BBH: $+5.0$; ARC-Challenge: $+3.7$; DROP: $+3.3$) and code/math domains (e.g., HumanEval: $+3.0$; MATH: $+2.4$; GSM8K: $+2.2$). 

Mechanistic analysis via LogitLens~\citep{nostalgebraist2020logitlens} and CKA~\citep{hendrycks2020measuring} reveals the source of these gains: Engram relieves the backbone from reconstructing static knowledge in early layers, thereby increasing effective depth available for complex reasoning.
Furthermore, by delegating local dependencies to lookups, Engram frees up attention capacity to focus on global context, enabling exceptional performance in long-context scenarios—substantially outperforming baselines on LongPPL~\citep{fangwrong} and RULER~\citep{hsiehruler} (e.g., Multi-Query NIAH: $97.0$ vs.\ $84.2$; Variable Tracking: $89.0$ vs.\ $77.0$).

Finally, we establish infrastructure-aware efficiency as a first-class principle. Unlike MoE's dynamic routing, Engram employs deterministic IDs to enable runtime prefetching, overlapping communication with computation. Empirical results show that offloading a 100B-parameter table to host memory incurs negligible overhead ($<3\%$). This demonstrates that Engram effectively bypasses GPU memory constraints, facilitating aggressive parameter expansion.

%% file: sections/arch.tex
\begin{figure}[htbp]
    \centering
    \includegraphics[width=0.95\linewidth]{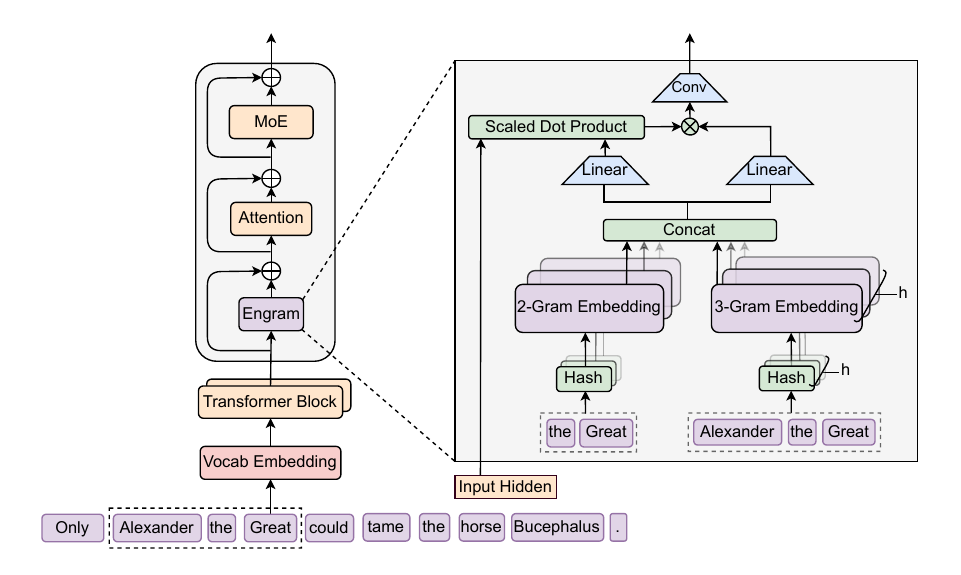}
    \caption{
        \textbf{The Engram Architecture.} The module augments the backbone by retrieving static $N$-gram memory and fusing it with dynamic hidden states via context-aware gating. This module is applied only to specific layers to decouple memory from compute, leaving the standard input embedding and un-embedding module intact.
    }
    \label{fig:dse_architecture}
\end{figure}

\section{Architecture}
\label{sec:architecture}

\subsection{Overview}
\label{sec:arch_overview}

As shown in \autoref{fig:dse_architecture}, Engram is a conditional memory module designed to augment the Transformer backbone by structurally separating static pattern storage from dynamic computation. Formally, given an input sequence $X = (x_1, \dots, x_T)$ and hidden states $\mathbf{H}^{(\ell)} \in \mathbb{R}^{T \times d}$ at layer $\ell$, the module processes each position $t$ in two functional phases: \textbf{retrieval and fusion}. 
First, as detailed in \autoref{sec:engram_retrieval}, we extract and compress suffix $N$-grams to deterministically retrieve static embedding vectors via hashing.
Subsequently, in \autoref{sec:engram_fusion}, these retrieved embeddings are dynamically modulated by the current hidden state and refined via a lightweight convolution.
Finally, we discuss the integration with multi-branch architectures in \autoref{sec:engram_hc_integration} and the system-level design in \autoref{sec:system_efficiency}.

\subsection{Sparse Retrieval via Hashed $N$-grams}
\label{sec:engram_retrieval}
The first phase maps local contexts to static memory entries, involving tokenizer compression and retrieving embeddings via deterministic hashing.

\paragraph{Tokenizer Compression}
While $N$-gram models typically operate directly on tokenizer outputs, standard subword tokenizers prioritize \textit{lossless reconstruction}, often assigning disjoint IDs to semantically equivalent terms~(e.g., \texttt{Apple} vs. \texttt{\textvisiblespace apple})~\citep{kudo2018sentencepiecesimplelanguageindependent,li2023starcodersourceyou}. To maximize semantic density, we implement a vocabulary projection layer. Specifically, we pre-compute a surjective function $\mathcal{P}: V \to V'$ that collapses raw token IDs into canonical identifiers based on normalized textual equivalence (using NFKC~\citep{UAX15-NFKC}, lowercasing, etc.). In practice, this process achieves a 23\% reduction in the effective vocabulary size for a 128k tokenizer (see ~\autoref{appendix:token_compression}). 
Formally, for a token at position $t$, we map its raw ID $x_t$ to a canonical ID $x'_t = \mathcal{P}(x_t)$ to form the suffix $N$-gram $g_{t,n} = (x'_{t-n+1}, \dots, x'_t)$.

\paragraph{Multi-Head Hashing.}

Directly parameterizing the combinatorial space of all possible $N$-grams is intractable. Following \citet{tito2017hash}, we adopt a hashing-based approach. To mitigate collisions, we employ $K$ distinct hash heads for each $N$-gram order $n$. Each head $k$ maps the compressed context to an index within an embedding table $\mathbf{E}_{n,k}$ (of prime size $M_{n,k}$) via a deterministic function $\phi_{n,k}$:
\begin{equation}
    z_{t,n,k} \triangleq \phi_{n,k}(g_{t,n}), \quad
    \mathbf{e}_{t,n,k} = \mathbf{E}_{n,k}[z_{t,n,k}].
\end{equation}
In practice, $\phi_{n,k}$ is implemented as a lightweight multiplicative-XOR hash. We construct the final memory vector $\mathbf{e}_t \in \mathbb{R}^{d_{\text{mem}}}$ by concatenating all retrieved embeddings:
\begin{equation}
    \mathbf{e}_t \triangleq \mathop{\Vert}_{n=2}^{N} \mathop{\Vert}_{k=1}^{K} \mathbf{e}_{t,n,k}.
\end{equation}

\subsection{Context-aware Gating}
\label{sec:engram_fusion}
The retrieved embeddings $\mathbf{e}_t$ serve as context-independent priors. Being static, however, they inherently lack contextual adaptability and may suffer from noise due to hash collisions or polysemy~\citep{haber-poesio-2024-polysemy}.
To enhance expressivity and resolve this ambiguity, we employ a context-aware gating mechanism inspired by Attention~\citep{vaswani2017attention,bahdanau2014neural}. Specifically, we utilize the current hidden state $\mathbf{h}_t$—which has aggregated global context via preceding attention layers—as a dynamic Query, while the retrieved memory $\mathbf{e}_t$ serves as the source for both Key and Value projections:
\begin{equation}
    \mathbf{k}_t = \mathbf{W}_K \mathbf{e}_t, \quad \mathbf{v}_t = \mathbf{W}_V \mathbf{e}_t
\end{equation}
where $\mathbf{W}_K, \mathbf{W}_V$ are learnable projection matrices. To ensure gradient stability~\citep{pmlr-v202-dehghani23a}, we apply RMSNorm~\citep{zhang2019root} to the Query and Key before computing the scalar gate $\alpha_t \in (0, 1)$: 
\begin{equation}
    \alpha_t = \sigma\left( \frac{\text{RMSNorm}(\mathbf{h}_t)^\top  \text{RMSNorm}(\mathbf{k}_t)}{\sqrt{d}} \right).
\end{equation}
The gated output is defined as $\tilde{\mathbf{v}}_t = \alpha_t \cdot \mathbf{v}_t$. This design enforces semantic alignment: if the retrieved memory $\mathbf{e}_t$ contradicts the current context $\mathbf{h}_t$, the gate $\alpha_t$ tends toward zero, effectively suppressing the noise.

Finally, to expand the receptive field and enhance the model's non-linearity, we introduce a short, depthwise causal convolution~\citep{gu2021efficiently,peng2023rwkv}. Let $\tilde{\mathbf{V}} \in \mathbb{R}^{T \times d}$ denote the sequence of gated values. Using a kernel size $w$ (set to 4), dilation $\delta$ (set to the max $N$-gram order) and SiLU activation~\citep{elfwing2018sigmoid}, the final output $\mathbf{Y}$ is computed as: 
\begin{equation}
    \mathbf{Y} = \text{SiLU}\left( \text{Conv1D}( \text{RMSNorm}(\tilde{\mathbf{V}}) ) \right) + \tilde{\mathbf{V}}, 
\end{equation}
The Engram module is integrated into the backbone via a residual connection: $\mathbf{H}^{(\ell)} \leftarrow \mathbf{H}^{(\ell)} + \mathbf{Y}$, followed by the standard Attention and MoE. 
Crucially, Engram is not applied to every layer; its specific placement is optimized to balance modeling effectiveness against the system-level latency constraints detailed in \autoref{sec:system_efficiency}.

\begin{figure}[!t]
    \centering
    \includegraphics[width=0.95\linewidth]{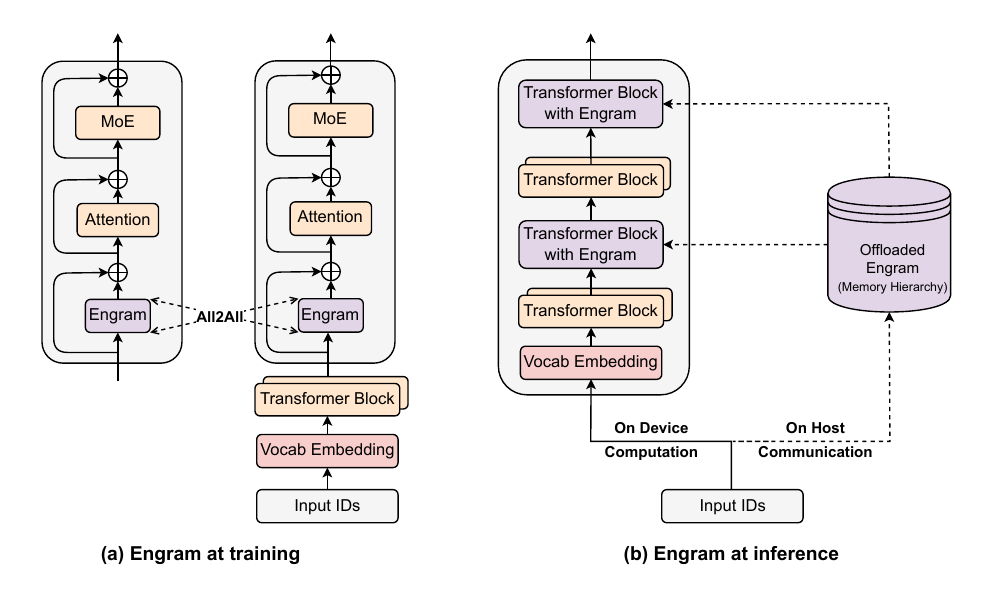}
    \caption{
    \textbf{System implementation of Engram.} 
    (a) Training Phase: The massive embedding tables are sharded across available GPUs. An All-to-All communication primitive is employed to retrieve active embedding rows across devices. 
    (b) Inference Phase: Engram tables are offloaded to host memory. By exploiting the deterministic retrieval logic, the host asynchronously prefetches and transfers embeddings, overlapping communication with the on-device computation of preceding Transformer blocks.
    }
    \label{fig:engram_system}
\end{figure}

\subsection{Integration with Multi-branch Architecture}
\label{sec:engram_hc_integration}

In this work, rather than standard single-stream connections~\citep{he2016deep}, we adopt the advanced multi-branch architecture as our default backbone, chosen for its superior modeling capabilities~\citep{larsson2016fractalnet,zhu2025hyperconnections,xie2025mhcmanifoldconstrainedhyperconnections,szegedy2015going}. A defining characteristic of this architecture is the expansion of the residual stream into $M$ parallel branches, where information flow is modulated by learnable connection weights.

Although the Engram module is inherently topology-agnostic, adapting it to this multi-branch framework necessitates structural optimization to balance efficiency and expressivity. 
Specifically, we implement a parameter-sharing strategy: a single sparse embedding table and a Value projection matrix $\mathbf{W}_V$ are shared across all $M$ branches, whereas $M$ distinct Key projection matrices $\{\mathbf{W}_K^{(m)}\}_{m=1}^M$ are employed to enable branch-specific gating behaviors.
For the $m$-th branch with hidden state $\mathbf{h}_t^{(m)}$, the branch-specific gating signal is computed as:
\begin{equation}
    \alpha_t^{(m)} = \sigma\left( \frac{\text{RMSNorm}(\mathbf{h}_t^{(m)})^\top \text{RMSNorm}(\mathbf{W}_K^{(m)} \mathbf{e}_t)}{\sqrt{d}} \right).
\end{equation}
The retrieved memory is then modulated by these independent gates applied to the shared value vector: $\mathbf{u}_t^{(m)} = \alpha_t^{(m)} \cdot (\mathbf{W}_V \mathbf{e}_t)$. This design allows the linear projections (one $\mathbf{W}_V$ and $M$ distinct $\mathbf{W}_K^{(m)}$) to be fused into a single dense FP8 matrix multiplication, maximizing the compute utilization of modern GPUs. Unless otherwise stated, all experiments utilize this integration with Manifold-Constrained Hyper-Connections~($M=4$)~\citep{xie2025mhcmanifoldconstrainedhyperconnections}.

\subsection{System Efficiency: Decoupling Compute and Memory}
\label{sec:system_efficiency}

Scaling model parameters is often constrained by the limited capacity of GPU high-bandwidth memory (HBM). However, unlike MoE which relies on runtime hidden states for dynamic routing, Engram employs a deterministic retrieval mechanism based solely on input token IDs, naturally decoupling parameter storage from computation. This predictability facilitates specialized optimization strategies for both training and inference, as illustrated in \autoref{fig:engram_system}.

During training, to accommodate large-scale embedding tables, we employ standard model parallelism by sharding the tables across available GPUs. An All-to-All communication primitive is used to gather active rows in the forward pass and dispatch gradients in the backward pass, enabling the total memory capacity to scale linearly with the number of accelerators.

During inference, this deterministic nature enables a prefetch-and-overlap strategy. Since memory indices are known prior to the forward pass, the system can asynchronously retrieve embeddings from abundant host memory via PCIe. To effectively mask communication latency, the Engram module is placed at specific layers within the backbone, leveraging the computation of preceding layers as a buffer to prevent GPU stalls.
This necessitates a hardware-algorithm co-design strategy: while placing Engram deeper extends the compute window available for hiding latency, our ablation in \autoref{sec:ablation_structural} shows that modeling performance favors early intervention to offload local pattern reconstruction. Therefore, the optimal placement must simultaneously satisfy both modeling and system latency constraints.

Furthermore, natural language $N$-grams inherently follow a Zipfian distribution~\citep{piantadosi2014zipf,Chao1950HumanBA}, where a small fraction of patterns accounts for the vast majority of memory accesses. This statistical property motivates a Multi-Level Cache Hierarchy: frequently accessed embeddings can be cached in faster storage tiers (e.g., GPU HBM or Host DRAM), while the long tail of rare patterns resides in slower, high-capacity media (e.g., NVMe SSD). This stratification allows Engram to scale to massive memory capacities with minimal impact on effective latency.

%% file: sections/scaling_law_exp.tex
\begin{figure}[t]
    \centering
    \includegraphics[width=0.95\linewidth]{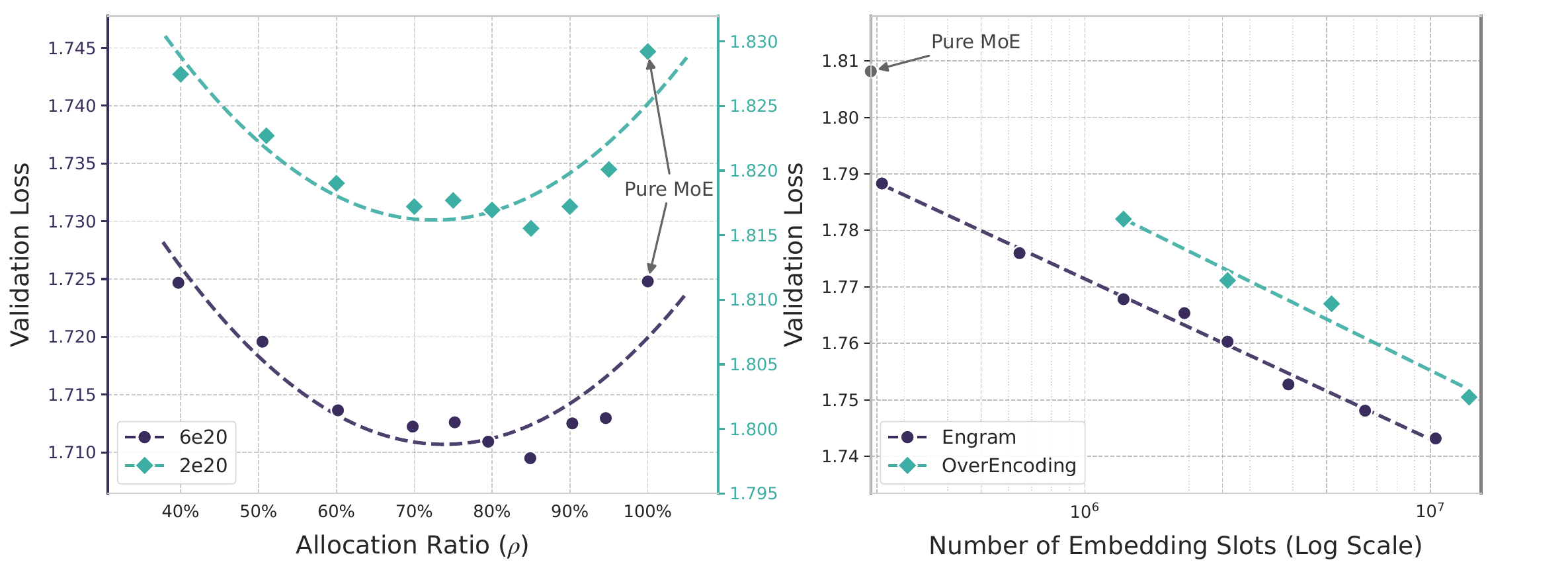}
    \caption{\textbf{Sparsity allocation and Engram scaling.}
        Left: Validation loss across allocation ratios $\rho$. Two compute budgets are shown ($2\text{e}20$ and $6\text{e}20$ FLOPs). Both regimes exhibit a U-shape, with hybrid allocation surpassing Pure MoE.
        Right: Scaling behavior in the infinite-memory regime. Validation loss exhibits a log-linear trend with respect to the number of embeddings.
    }
    \label{fig:scaling_law}
\end{figure}

\section{Scaling Laws and Sparsity Allocation}
\label{sec:scaling_law}

Engram, as an instantiation of conditional memory, is structurally complementary to the conditional computation provided by MoE experts.
This section investigates the scaling properties of this duality and how to optimally allocate sparse capacity. Specifically, two key questions drive our research:
\begin{enumerate}[leftmargin=*,itemsep=2pt,topsep=2pt]
    \item \textbf{Allocation under Finite Constraints.} When total parameters and training compute are fixed (Iso-parameter and Iso-FLOPs), how should we split the sparse capacity between MoE experts and Engram embeddings?
    \item \textbf{Infinite Memory Regime.} Considering the non-scaling $\mathcal{O}(1)$ overhead of Engram, if the memory budget is relaxed or scaled aggressively, what scaling behavior does Engram exhibit by itself?
\end{enumerate}

\subsection{Optimal Allocation Ratio Between MoE and Engram}
\label{sec:ratio_scaling_law}

\paragraph{Compute-matched formulation.}
We analyze the trade-off using three parameter metrics:
\begin{itemize}[leftmargin=*,itemsep=2pt,topsep=2pt]
    \item $P_{\mathrm{tot}}$: total trainable parameters, excluding vocabulary embedding and LM head.
    \item $P_{\mathrm{act}}$: activated parameters per token. This quantity determines the training cost (FLOPs).
    \item $P_{\mathrm{sparse}} \triangleq P_{\mathrm{tot}} - P_{\mathrm{act}}$: the \textit{inactive} parameters, which represents the ``free'' parameter budget available for scaling model size without incurring computational cost~(e.g., unselected experts or unretrieved embeddings).
\end{itemize}
We keep $P_{\mathrm{tot}}$ and $P_{\mathrm{act}}$ fixed within each FLOPs budget, so that models have the same number of parameters and the same per-token FLOPs. For MoE, $P_{\mathrm{act}}$ is determined by the top-$k$ selected experts, while the parameters of non-selected experts contribute to $P_{\mathrm{sparse}}$. For Engram, only a constant number of slots are retrieved per token, so scaling the number of embedding slots increases $P_{\mathrm{tot}}$ without increasing per-token FLOPs.

\paragraph{Allocation ratio.}
We define the allocation ratio $\rho\in[0,1]$ as the fraction of the inactive-parameter budget assigned to MoE expert capacity:
\begin{equation}
P_{\mathrm{MoE}}^{(\mathrm{sparse
})} = \rho\, P_{\mathrm{sparse}}, 
\qquad
P_{\mathrm{Engram}} = (1-\rho)\, P_{\mathrm{sparse}}.
\end{equation}
Intuitively:
\begin{itemize}[leftmargin=*,itemsep=2pt,topsep=2pt]
    \item $\rho=1$ corresponds to a pure MoE model (all inactive parameters are routed experts).
    \item $\rho<1$ reduces the number of routed experts and reallocates the freed parameters to Engram embedding slots.
\end{itemize}

\paragraph{Experimental protocol.}
We evaluate this trade-off at two compute regimes and maintain a constant sparsity ratio $P_{\mathrm{tot}} / P_{\mathrm{act}} \approx 10$ across both settings:
\begin{itemize}[leftmargin=*,itemsep=2pt,topsep=2pt]
    \item \textbf{$C=2\times 10^{20}$} FLOPs: $P_{\mathrm{tot}}\approx 5.7$B and $P_{\mathrm{act}}=568$M. The baseline ($\rho=1$) has a total of 106 experts.
    \item \textbf{$C=6\times 10^{20}$} FLOPs: $P_{\mathrm{tot}}\approx 9.9$B and $P_{\mathrm{act}}=993$M. The baseline ($\rho=1$) has a total of 99 experts.
\end{itemize}
For different $\rho$, we construct the corresponding model only by adjusting the number of routed experts and the number of Engram embedding slots.
All runs use the identical training pipeline and optimization hyperparameters.

\paragraph{Results and Analysis.}
\autoref{fig:scaling_law} (left) reveals a consistent U-shaped relationship between validation loss and the allocation ratio $\rho$.
Remarkably, the Engram model achieves comparable performance to the pure MoE baseline ($\rho=100\%$) even when the MoE allocation is reduced to just $\rho \approx 40\%$ (i.e., a total of 46 experts for the 5.7B model and 43 experts for the 9.9B model).
Furthermore, the pure MoE baseline proves suboptimal: reallocating roughly $20\%\text{--}25\%$ of the sparse parameter budget to Engram yields the best performance.
Quantitatively, in the 10B regime ($C=6\times 10^{20}$), validation loss improves from $1.7248$ (at $\rho=100\%$) to $1.7109$ near the optimum of $\rho\approx 80\%$ ($\Delta=0.0139$).
Crucially, the location of this optimum is stable across regimes ($\rho \approx 75\%\text{--}80\%$), suggesting a robust allocation preference across the examined scales (under fixed sparsity). This observed U-shape confirms the structural complementarity between the two modules:
\begin{itemize}[leftmargin=*,itemsep=2pt,topsep=2pt]
    \item \textbf{MoE-dominated ($\rho \to 100\%$):} The model lacks dedicated memory for static patterns, forcing it to inefficiently reconstruct them through depth and computation.
    \item \textbf{Engram-dominated ($\rho \to 0\%$):} The model loses conditional computation capacity, hurting tasks that require dynamic, context-dependent reasoning; memory cannot replace computation in this regime.
\end{itemize}

\subsection{Engram under Infinite Memory Regime}
\label{sec:infinite_memory}

In \autoref{sec:ratio_scaling_law}, we optimized the allocation under a fixed parameter budget. We now explore the complementary setting: aggressive memory scaling.
This investigation is motivated by Engram's unique ability to decouple storage from compute detailed in \autoref{sec:system_efficiency}.

\paragraph{Experimental protocol.}
We utilize a fixed MoE backbone with $P_{\mathrm{tot}}\approx 3$B and $P_{\mathrm{act}}=568$M, trained for 100B tokens to ensure convergence.
On top of this backbone, we attach an Engram table and sweep the number of slots $M$ from $2.58\times 10^5$ to $1.0\times 10^7$ (adding up to $\approx 13$ billion parameters).
For baselines, we compare against OverEncoding~\citep{huang2025over}, which integrates $N$-gram embeddings via averaging with the vocabulary embedding.
We note that while other work such as SCONE~\citep{yu2025scaling} also investigates large-scale embeddings, it is primarily inference-focused and includes extra module~(\textit{f-gram model}) and additional training FLOPs, rendering it incompatible with the strict iso-compute constraints of this study.

\paragraph{Results.}
\autoref{fig:scaling_law} (right) demonstrates that scaling the number of memory slots yields a clear and consistent improvement in validation loss.
Across the explored range, the curve follows a strict power law (linear in log-space), indicating that Engram provides a predictable scaling knob: larger memory continues to pay off without requiring additional computation.
Crucially, regarding scaling efficiency: while the direct averaging approach of OverEncoding benefits from larger memory tables, Engram unlocks much larger scaling potential from the same memory budget. Together with the allocation law in \autoref{sec:ratio_scaling_law}, these results validate that conditional memory serves as a distinct, scalable axis of sparse capacity that complements the conditional computation of MoE.

%% file: sections/large_scale_exp.tex
\section{Large Scale Pre-training}
\definecolor{refgray}{gray}{0.5}
\begin{table}[!h]
    \centering
        \caption{\textbf{Pre-training performance comparison between dense, MoE, and Engram models}. All models are trained for 262B tokens and are matched in activated parameters (3.8B). Engram-27B is iso-parameters with MoE-27B by reallocating parameters from routed experts~(72 $\to$ 55) to a 5.7B-parameter Engram memory. Engram-40B further increases Engram memory (18.5B parameters) while keeping the activated-parameter budget fixed. Full training-time benchmark trajectories are reported in \autoref{appendix:benchmark_curve}.}
    \footnotesize
    \setlength{\tabcolsep}{4.5pt}
     \begin{tabular}{@{}c l c | c | c c !{\color{refgray}\vrule} >{\color{refgray}}c@{}}
    \toprule
        & \textbf{Benchmark \tiny(Metric)} & \textbf{\# Shots} & \textbf{Dense-4B} & \textbf{MoE-27B} & \textbf{Engram-27B} & \textbf{\color{refgray}Engram-40B} \\
    \midrule
    & \# Total Params &  & 4.1B & 26.7B & 26.7B & 39.5B \\
    & \# Activated  \tiny(w/o token embed) &  & 3.8B & 3.8B & 3.8B & 3.8B \\
    & \# Trained Tokens &  & 262B & 262B & 262B &  262B \\
    & \# Experts \tiny(shared + routed, top-$k$) &  & - & $2+72$ (top-6) & $2+55$ (top-6) & $2+55$ (top-6) \\
    & \# Engram Params &  & - & - & 5.7B & 18.5B \\
    \midrule
    \multirow{2}{*}{\shortstack{Language\\Modeling}} & Pile {\tiny (loss)} & - & 2.091 & 1.960 & \textbf{1.950} & 1.942 \\
    & Validation Set {\tiny (loss)} & - & 1.768 & 1.634 & \textbf{1.622} & 1.610 \\
    \midrule
    \multirow{15}{*}{\shortstack{Knowledge\\\&\\Reasoning}} 
    & MMLU {\tiny (Acc.)} & 5-shot & 48.6 & 57.4 & \textbf{60.4} &  60.6 \\
    & MMLU-Redux {\tiny (Acc.)} & 5-shot & 50.7 & 60.6 & \textbf{64.0} & 64.5 \\
    & MMLU-Pro {\tiny (Acc.)} & 5-shot & 21.1 & 28.3 & \textbf{30.1} & 31.3  \\
    & CMMLU {\tiny (Acc.)} & 5-shot & 47.9 & 57.9 & \textbf{61.9} &  63.4 \\
    & C-Eval {\tiny (Acc.)} & 5-shot & 46.9 & 58.0 & \textbf{62.7} & 63.3 \\
    & AGIEval {\tiny (Acc.)} & 0-shot & 29.1 & 38.6 & \textbf{41.8} & 45.9 \\
    & ARC-Easy {\tiny (Acc.)} & 25-shot & 76.8 & 86.5 & \textbf{89.0} & 90.1 \\
    & ARC-Challenge {\tiny (Acc.)} & 25-shot & 59.3 & 70.1 & \textbf{73.8} &  76.4 \\
    & TriviaQA {\tiny (EM)} & 5-shot & 33.0 & 48.8 & \textbf{50.7} & 51.8 \\
    & TriviaQA-ZH {\tiny (EM)} & 5-shot & 62.8 & 74.8 & \textbf{76.3} & 77.9 \\
    & PopQA {\tiny (EM)} & 15-shot & 15.1 & 19.2 & \textbf{19.4} & 21.2 \\
    & CCPM {\tiny (Acc.)} & 0-shot & 72.2 & 79.6 & \textbf{87.1} &  87.7 \\
    & BBH {\tiny (EM)} & 3-shot & 42.8 & 50.9 & \textbf{55.9} &  57.5 \\
    & HellaSwag {\tiny (Acc.)} & 0-shot & 64.3 & 71.8 & \textbf{72.7} & 73.1 \\
    & PIQA {\tiny (Acc.)} & 0-shot & 63.8 & 71.9 & \textbf{73.5} &  76.5 \\
    & WinoGrande {\tiny (Acc.)} & 5-shot & 64.0 & 67.6 & \textbf{67.8} &  68.1 \\
    \midrule
    \multirow{4}{*}{\shortstack{Reading\\Comprehension}} 
    & DROP {\tiny (F1)} & 1-shot & 41.6 & 55.7 & \textbf{59.0} & 60.7 \\
    & RACE-Middle {\tiny (Acc.)} & 5-shot & 72.4 & 80.9 & \textbf{82.8} & 83.3 \\
    & RACE-High {\tiny (Acc.)} & 5-shot & 66.0 & 75.4 & \textbf{78.2} & 79.2 \\
    & C3 {\tiny (Acc.)} & 0-shot & 57.7 & 60.1 & \textbf{63.6} &  61.8 \\
    \midrule
    \multirow{8}{*}{\shortstack{Code \& Math}} 
    & HumanEval {\tiny (Pass@1)} & 0-shot & 26.8 & 37.8 & \textbf{40.8} & 38.4  \\
    & MBPP {\tiny (Pass@1)} & 3-shot & 35.4 & 46.6 & \textbf{48.2} & 46.2 \\
    & CruxEval-i {\tiny (EM)} & 0-shot &27.6&30.7 & \textbf{32.2} & 36.2 \\
    & CruxEval-o {\tiny (EM)} & 0-shot &28.7&34.1 & \textbf{35.0} & 35.3 \\
    & GSM8K {\tiny (EM)} & 8-shot & 35.5 & 58.4 & \textbf{60.6} & 62.6 \\
    & MGSM {\tiny (EM)} & 8-shot & 27.0 & 46.8 & \textbf{49.4} &  52.4 \\
    & MATH {\tiny (EM)} & 4-shot & 15.2 & 28.3 & \textbf{30.7} & 30.6 \\
    \bottomrule
    \end{tabular}

    \label{tab:27b_eval}
\end{table}

With the proposed Engram architecture and the empirically derived allocation law, we scale Engram to the multi-billion parameter to validate its efficacy in real-world language model pre-training. Specifically, we train four models:
(1) \textbf{Dense-4B} (4.1B total parameters),
(2) \textbf{MoE-27B} (26.7B total parameters),
(3) \textbf{Engram-27B} (26.7B total parameters), and
(4) \textbf{Engram-40B} (39.5B total parameters).
All models are trained using an identical data curriculum (same token budget and order) and are strictly matched in the number of activated parameters.

\subsection{Experimental Setup}

\paragraph{Training Data and Model Configurations} All models are pre-trained on a corpus of 262 billion tokens and we utilize the tokenizer from DeepSeek-v3~\citep{liu2024deepseek} with a vocabulary size of 128k. For modeling, to ensure a controlled comparison, we adhere to a consistent default setting across all models unless explicitly stated otherwise. We utilize a 30-block Transformer with a hidden size of 2560. Each block integrates a Multi-head Latent Attention (MLA)~\citep{deepseekai2024deepseekv2strongeconomicalefficient} with 32 heads, connected to FFNs via mHC~\citep{xie2025mhcmanifoldconstrainedhyperconnections} with an expansion rate of 4. All models are optimized using Muon~\citep{jordan2024muon,liu2025muon}; detailed hyperparameters are listed in the \autoref{appendix:detailed_model_arch}. We instantiate four distinct models:

\begin{itemize}
    \item \textbf{Dense-4B} serves as the baseline model. It utilizes the backbone architecture described above, incorporating a standard dense FFN into every block.

    \item \textbf{MoE-27B} replaces the standard dense FFN with a DeepSeekMoE module~\citep{dai2024deepseekmoe}. Configured with 72 routed experts and 2 shared experts (activating the top-$k=6$ routed experts per token), this model scales to 26.7B total parameters while maintaining the same activated parameters as Dense-4B.

    \item \textbf{Engram-27B} is strictly derived from the \textbf{MoE-27B} architecture to ensure fair comparison. We reduce the number of routed experts from 72 to 55 and reallocate the freed parameters to a 5.7B-parameter embedding module~($\rho=74.3\%$), keeping the total model size constant at 26.7B. Regarding the Engram configuration, we instantiate the module at layers 2 and 15 and set the maximum $N$-gram size to 3, the number of heads to 8, and the dimension to 1280. For optimization, the embedding parameters are updated using Adam~\citep{kingma2014adam} with a learning rate scaled by $5\times$ and no weight decay, while the convolution parameters are initialized to zero to strictly preserve the identity mapping at the start of training.

    \item \textbf{Engram-40B} retains the same backbone and computation budget as Engram-27B but scales the sparse embedding module to 18.5B parameters (totaling 39.5B parameters). This model is designed to investigate the scaling properties of Engram.
\end{itemize}

\paragraph{Evaluation Protocol}
We evaluate models on a diverse suite of benchmarks spanning language modeling, knowledge, reasoning, reading comprehension, and code/math.
For each benchmark, we follow standard prompting protocols and evaluation metrics.

\begin{itemize}
    \item \textbf{Language Modeling:}
    We report loss on the test set of The Pile~\citep{gao2020pile} and an validation set drawn from the same distribution as the training data.

    \item \textbf{Knowledge \& Reasoning:}
    MMLU~\citep{hendrycks2020measuring}, MMLU-Redux~\citep{gema2025we}, MMLU-Pro~\citep{wang2024mmlu}, CMMLU~\citep{li2024cmmlu}, C-Eval~\citep{huang2023c}, AGIEval~\citep{zhong2024agieval}, ARC-Easy/Challenge~\citep{clark2018think}, TriviaQA~\citep{joshi2017triviaqa}, TriviaQA-ZH (internal), PopQA~\citep{mallen2023not}, CCPM~\citep{li2021ccpm}, BBH~\citep{suzgun2023challenging}, HellaSwag~\citep{zellers2019hellaswag}, PIQA~\citep{bisk2020piqa}, and WinoGrande~\citep{sakaguchi2021winogrande}.

    \item \textbf{Reading Comprehension:}
    DROP~\citep{dua2019drop}, RACE (Middle/High)~\citep{lai2017race}, and C3~\citep{sun2020investigating}.

    \item \textbf{Code \& Math:}
    HumanEval~\citep{chen2021evaluatinglargelanguagemodels}, MBPP~\citep{austin2021program}, CruxEval~\citep{gu2024cruxeval}, GSM8K~\citep{cobbe2021training}, MGSM~\citep{shi2022language}, and MATH~\citep{hendrycks2021measuring}.
\end{itemize}

\subsection{Experimental Results}

\autoref{tab:27b_eval} summarizes the main results.
First, consistent with prior literature~\citep{shazeer2017outrageously,he2024mixture,borgeaud2022improving}, sparse architectures demonstrate superior scaling laws compared to dense models. Under the same training compute budget, all three sparse variants (MoE-27B, Engram-27B/40B) significantly outperform the iso-FLOPs Dense-4B baseline across all benchmarks.

More importantly, Engram-27B consistently improves over the iso-parameter and iso-FLOPs MoE-27B baseline. Interestingly, these gains are not limited to knowledge-intensive tasks (e.g., MMLU: +3.0, MMLU-Pro: +1.8, CMMLU: +4.0), where memory capacity is intuitively beneficial. We observe even more significant improvements in general-reasoning domains (e.g., BBH: +5.0, ARC-Challenge: +3.7, DROP: +3.3), as well as code and mathematical reasoning (e.g., HumanEval: +3.0, MBPP: +1.6, GSM8K: +2.2, MATH: +2.4). To reduce the impact of benchmark noise and to visualize training dynamics, we provide full benchmark trajectories during pre-training in \autoref{appendix:benchmark_curve}.
These results support our hypothesis that introducing a dedicated knowledge lookup primitive improves representation efficiency beyond what can be achieved by allocating the entire sparse budget to conditional computation.

Finally, scaling to Engram-40B further reduces pre-training loss and improves performance across most benchmarks. Although it does not yet strictly dominate Engram-27B on every task, this is likely an artifact of under-training. We observe that the training loss gap between Engram-40B and the baselines continues to widen towards the end of training, suggesting that the expanded memory capacity has not yet fully saturated within the current token budget.

%% file: sections/long_context_exp.tex
\begin{table}[t]
\centering
\caption{
\textbf{Long-context performance comparison.} Parenthetical values (e.g. \textit{(50k, 1.62)}) denote the pre-training steps and the corresponding loss prior to the long-context extension. Two key findings: (1) With only 82\% of the pre-training FLOPs (41k vs. 50k), Engram-27B matches the baseline's LongPPL~\citep{fangwrong} performance while achieving significantly higher accuracy on RULER~\citep{hsiehruler}; (2) Under both iso-pretraining-loss (46k) and iso-pretraining-FLOPs (50k) settings, Engram-27B substantially outperforms the baseline across all metrics. 
\textbf{Bold} indicates the best and \underline{underline} the second.
}

\resizebox{0.95\linewidth}{!}{%
\begin{tabular}{@{}l cccc cccc cccc@{}}
\toprule
\multirow{3}{*}{\textbf{Model}} & \multicolumn{4}{c}{\textbf{LongPPL (32k)}} & \multicolumn{8}{c}{\textbf{RULER (32k)}} \\
\cmidrule(lr){2-5} \cmidrule(l){6-13}
 & \multicolumn{4}{c}{Perplexity ($\downarrow$)} & \multicolumn{4}{c}{NIAH Accuracy ($\uparrow$)} & \multicolumn{4}{c}{Other Tasks ($\uparrow$)} \\
\cmidrule(lr){2-5} \cmidrule(lr){6-9} \cmidrule(l){10-13}
 & Book & Paper & Code & L-CoT & S & MK & MV & MQ & VT & CWE & FWE & QA \\ \midrule

MoE-27B \textit{(50k, 1.63)} & 4.38 & 2.91 & 2.49 & 14.16 & \textbf{100.0} & 88.0 & 92.7 & 84.2 & 77.0 & \underline{4.5} & 73.0 & 34.5 \\ 
\midrule
Engram-27B \textit{(41k, 1.66)} & 4.37 & 2.92 & 2.50 & 14.26 & \underline{99.6} & 88.3 & 93.0 & 89.5 & 83.2 & 3.8 & 99.6 & \textbf{44.0}\\
Engram-27B \textit{(46k, 1.63)} & \underline{4.19} & \underline{2.84} & \underline{2.45} & \underline{13.59} & 97.6 & \underline{89.0} & \underline{95.5} & \textbf{97.0} & \underline{87.2} & 4.3 & \underline{98.6} & 37.5 \\
Engram-27B \textit{(50k, 1.62)} & \textbf{4.14} & \textbf{2.82} & \textbf{2.44} & \textbf{13.41} & 99.3 & \textbf{89.3} & \textbf{96.5} & \textbf{97.0} & \textbf{89.0} & \textbf{5.9} & \textbf{99.3} & \underline{40.5} \\ \bottomrule
\end{tabular}%
}

\label{tab:long_eval}
\end{table}

\section{Long Context Training}
\label{sec:long_ctx}

By offloading local dependency modeling to static lookups, the Engram architecture preserves valuable attention capacity for managing global context. In this section, we empirically verify this structural advantage by conducting long-context extension training~\citep{gao2025train,peng2023yarn}. Through a rigorous evaluation protocol that isolates architectural contributions from base model capabilities, we demonstrate that Engram yields significant gains in long-range retrieval and reasoning tasks.

\subsection{Experimental Setup}

\paragraph{Training Details.} To enable long-context capabilities, we adopt the context expansion strategy introduced in DeepSeek-V3~\citep{liu2024deepseek}. Following the pre-training stage, we apply YaRN~\citep{peng2023yarn} for context window extension in a 32768-token context training stage for 5,000 steps~(30B tokens of high-quality, long-context data). The hyper-parameters are scale $s=10, \alpha=1,\beta=32$ and the scaling factor $f=0.707$.

\paragraph{Model Configurations.} We compare context extensions across four distinct model configurations. We utilize the final pre-training checkpoints (50k steps) for both MoE-27B and Engram-27B. Additionally, to rigorously benchmark architectural efficiency, we select two intermediate checkpoints for Engram-27B at 41k and 46k steps. 
Despite differing initialization stages, all variants undergo \textit{the exact same} context extension training protocol.
Crucially, Engram-27B (46k) is selected because it exhibits the same pre-training loss as the fully trained MoE-27B (50k). This creates a controlled "Iso-Loss" setting, ensuring that any performance divergence during context extension is attributable to the architecture rather than the starting quality of the model.

\paragraph{Evaluation Benchmarks.} We assess long-context performance using LongPPL~\citep{fangwrong} and RULER~\citep{hsiehruler}. For LongPPL, we construct evaluation sets spanning four categories: long books, research papers, code repositories, and long chain-of-thought (CoT) trajectories. For RULER, we evaluate on 14 subsets aggregated into 8 categories: Single (S), Multi-keys (MK), Multi-values (MV) and Multi-queries (MQ) Needle-in-a-Haystack; Multi-hop Variable Tracking (VT), Common Words Extraction (CWE), Frequent Words Extraction (FWE), and Question Answering (QA).

\subsection{Experimental Results}

The evaluation results are summarized in~\autoref{tab:long_eval}. To accurately assess the contribution of the Engram architecture, our analysis proceeds in two steps: first, decoupling the impact of base model capability from architectural design, and second, conducting a controlled analysis.

\textbf{1. Long-Context Capability Beyond Attention Mechanics.} 
While attention mechanisms and positional encoding provide the structural basis for context processing~\citep{su2023roformerenhancedtransformerrotary,press2021train,yang2025path,xiao2023efficient}, our results indicate that long-context performance is not solely determined by architectural priors. Observing the trajectory of Engram (41k $\rightarrow$ 50k), we find that long-context performance improves monotonically with pre-training progression, even when controlling for identical model architecture and a fixed computational budget during the context extension stage.
This suggests that long-context performance is intrinsically coupled with the general modeling ability of the base model. Consequently, a rigorous architectural comparison must control for this confounding variable by aligning base model loss, rather than merely aligning training steps.

\textbf{2. Architectural Superiority under Controlled Settings.} 
Guided by the principle above, we benchmark Engram against the MoE baseline. When controlling for base capability, the efficiency gains of the Engram module become evident:
\begin{itemize}
    \item \textbf{Iso-Loss Setting (46k vs.\ Baseline):} This setting strictly isolates architectural efficiency. When comparing Engram-27B (46k) against the fully trained MoE-27B (50k)—models aligned on pre-training loss—Engram demonstrates significant gains. Specifically, it outperforms the baseline on complex retrieval tasks (e.g., Multi-Query NIAH: $97.0$ vs.\ $84.2$; VT: $87.2$ vs.\ $77.0$). 
    \item \textbf{Iso-FLOPs Setting (50k vs.\ Baseline):} Under the standard iso-compute budget, Engram-27B (50k) further widens this gap, establishing the highest performance across the board.
    \item \textbf{Extreme Setting ($\approx82\%$ Compute):} Even the early-stopped Engram-27B (41k) remains highly competitive against the fully trained MoE-27B (50k). It matches the baseline on LongPPL and surpasses it on RULER, underscoring the intrinsic superiority of the Engram architecture.
\end{itemize}

%% file: sections/analysis.tex
\begin{figure}[t]
    \centering
    \includegraphics[width=\linewidth]{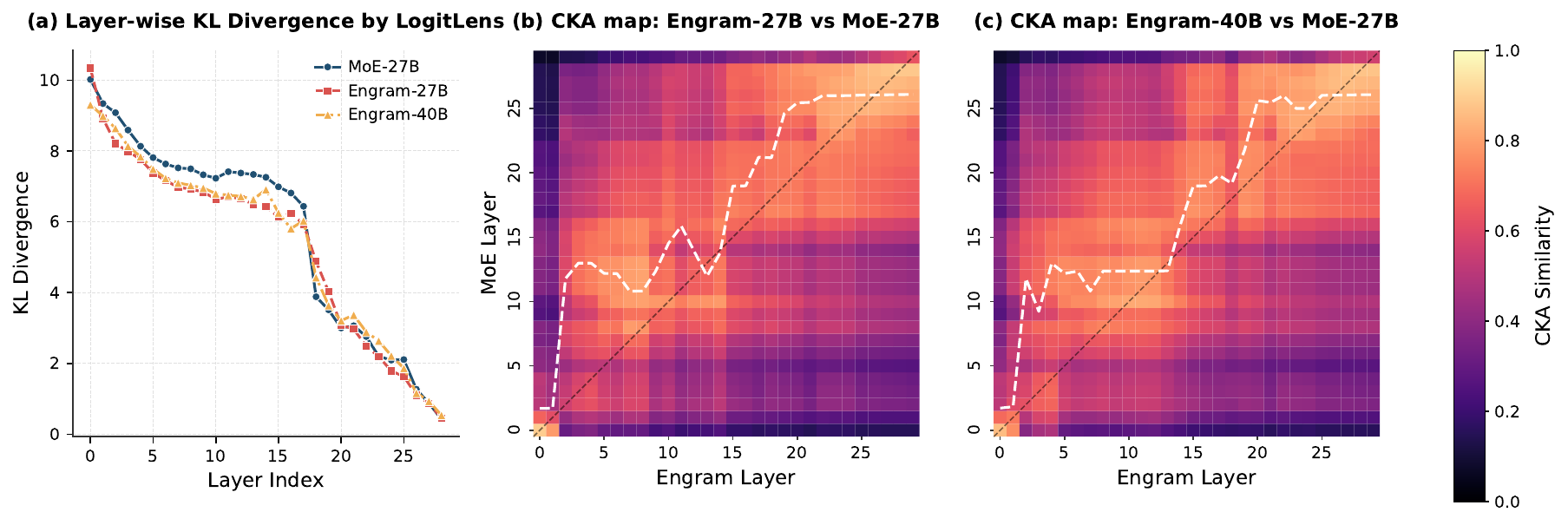}
    \caption{\textbf{Analysis of representational alignment and convergence speed.} (a) Layer-wise KL Divergence via LogitLens~\citep{nostalgebraist2020logitlens}. The consistently lower divergence in early layers indicates that Engram accelerates prediction convergence. (b-c) Similarity heatmap computed by CKA~\citep{kornblith2019similarity}. The distinct upward shift of the high-similarity diagonal demonstrates that =Engram's shallow layers are functionally equivalent to deeper layers of the MoE model, effectively increasing the model's depth.}
    \label{fig:logitlens_and_cka}
\end{figure}

\begin{table}[htbp]
    \centering
    \small
    \renewcommand{\arraystretch}{1.4}
    \caption{
    \textbf{Entity resolution example reproduced from} \citet{ghandeharioun2024patchscopes}. 
    This table illustrates how LLMs gradually integrate context tokens through layers of attention and FFNs to construct the internal representation of the entity: \texttt{``Diana, Princess of Wales''}. The ``Latent State Translation'' column displays the automatically generated text for the last token: \texttt{``Wales''} by PatchScope~\citep{ghandeharioun2024patchscopes}, while the ``Explanation'' column presents the manual interpretation provided by the original authors.
    }

    \begin{tabularx}{0.85\textwidth}{ 
        c 
        >{\RaggedRight\arraybackslash}X 
        >{\RaggedRight\arraybackslash}m{4.5cm} 
    }
        \toprule
        \textbf{Layer} & \textbf{Latent State Translation} & \textbf{Explanation} \\
        \midrule
        
        1-2 &
        : Country in the United Kingdom &
        \textbf{Wales} \\
        
        3 &
        : Country in Europe &
        \textbf{Wales} \\
        
        4 &
        : Title held by female sovereigns in their own right or by queens consort &
        \textbf{Princess of Wales} \newline (unspecific) \\
        
        5 &
        : Title given to the wife of the Prince of Wales (and later King) &
        \textbf{Princess of Wales} \newline (unspecific) \\
        
        6 &
        : Diana, Princess of Wales (1961-1997), the first wife of Prince Charles, Prince of Wales, who was famous for her beauty and humanitarian work &
        \textbf{Diana, \newline Princess of Wales} \\
        \bottomrule
    \end{tabularx}
    \label{tab:patchscope}
\end{table}

\section{Analysis}
In this section, we investigate the internal mechanisms of Engram, including its effective depth~(\autoref{sec:depth_analysis}), core module design~(\autoref{sec:ablation_structural}), and parametric sensitivity~(\autoref{sec:dse_sensitive}). Additionally, we evaluate the inference throughput with offloading~(\autoref{sec:inference_analysis}) and conclude with a case study~(\autoref{sec:case_study}).

\subsection{Is Engram functionally equivalent to increasing the model’s depth?}
\label{sec:depth_analysis}

Current LLMs lack a dedicated knowledge lookup primitive and they rely on computation to simulate memory recall. As shown in \autoref{tab:patchscope}, to recognize the entity $\texttt{"Diana,\,Princess\;of\;Wales"}$, an LLM must consume multiple layers of Attention and FFNs to progressively compose features~\citep{li2025echoesbertmodernlanguage,ghandeharioun2024patchscopes,DBLP:conf/coling/JinYHZWH0MMDYDZ25}, a process that could theoretically be identified via a knowledge lookup operation.

Given this, we posit that by equipping the model with an explicit knowledge lookup capability, Engram effectively mimics an increase in model depth by relieving the model of the early stages of feature composition. To validate this hypothesis, we employ two mechanistic interpretability tools: LogitLens~\citep{nostalgebraist2020logitlens,belrose2023eliciting} and Centered Kernel Alignment analysis (CKA)~\citep{kornblith2019similarity,davari2022reliability}.

\subsubsection{Accelerated Prediction Convergence}
\label{sec:logitlens}

We first analyze the evolution of predictions across layers using LogitLens~\citep{nostalgebraist2020logitlens}. By projecting each intermediate layer's hidden state with the final LM Head, we compute the Kullback–Leibler divergence~\citep{kullback1951information} between the intermediate output distribution and the model's final output distribution. This metric quantifies how close a latent representation is to being ``prediction-ready''~\citep{csordas2025language,belrose2023eliciting}.

\autoref{fig:logitlens_and_cka} (a) reports the layer-wise KL divergence. Compared to the MoE baseline, both Engram variants exhibit systematically smaller KL divergence, with the most pronounced gap appearing in the early blocks. The steeper descent in the Engram curves indicates that the model finishes feature composition much faster. This observation aligns with our hypothesis: by accessing external knowledge explicitly, Engram reduces the computational steps required, thereby reaching high-confidence, valid predictions earlier in the network hierarchy.

\subsubsection{Representational Alignment and Effective Depth}

To further investigate whether Engram layers semantically correspond to deeper layers of the baseline, we employ Centered Kernel Alignment~(CKA), a widely established metric for comparing representational structures~\citep{kornblith2019similarity,kriegeskorte2008representational}. Given two sets of representations $X$ and $Y$ (e.g., activations from different models or layers), CKA is defined as:
\begin{equation}
    \text{CKA}(K, L) = \frac{\text{HSIC}(K, L)}{\sqrt{\text{HSIC}(K, K)\text{HSIC}(L, L)}}
\end{equation}
where $K = XX^\top$ and $L = YY^\top$ denote the Gram matrices (using a linear kernel) and HSIC is Hilbert-Schmidt Independence Criterion~\citep{gretton2005measuring}. We employ a minibatch implementation with an unbiased estimator of HSIC~\citep{davari2022reliability} and evaluate on the Few-NERD dataset~\citep{ding2021few}, extracting hidden states corresponding to the final token of named entities.

To rigorously quantify the layer-wise correspondence, 
we first compute the pairwise CKA similarity matrix $S \in [0, 1]^{L \times L}$, where $L$ is the number of layers. We then introduce a soft alignment index $a_j$, defined as the weighted centroid of the top-$k$ most similar MoE layers for each Engram layer $j$:
\begin{equation}
    a_j = \frac{\sum_{i \in \mathcal{I}_j} S_{i,j} \cdot i}{\sum_{i \in \mathcal{I}_j} S_{i,j}}, \quad \text{where } \mathcal{I}_j = \mathop{\text{argtop}k}_{i} (S_{i,j}). 
\end{equation}
Here, $S_{i,j}$ denotes the similarity score between MoE layer $i$ and Engram layer $j$. The index $a_j$ serves as a robust proxy for the ``effective MoE depth'' corresponding to Engram layer $j$, utilizing top-$k$ filtering (with $k=5$) to mitigate low-similarity noise.

\autoref{fig:logitlens_and_cka} (b)--(c) visualize the similarity heatmaps overlayed with the soft alignment curve (dashed white line). We observe a distinct upward shift from the diagonal, meaning that $a_j > j$ for a wide range of layers. For instance, the representations formed at layer 5 of Engram-27B align most closely with those at approximately layer 12 of the MoE baseline.

The consistent off-diagonal shift, which aligns with the LogitLens results~(\autoref{sec:logitlens}), confirms that Engram achieves deeper representations at earlier layers. This validates our central hypothesis: by bypassing early-stage feature composition via explicit lookups, Engram is functionally equivalent to increasing the model's effective depth.

\begin{figure}[t]
\centering
\includegraphics[width=0.75\linewidth]{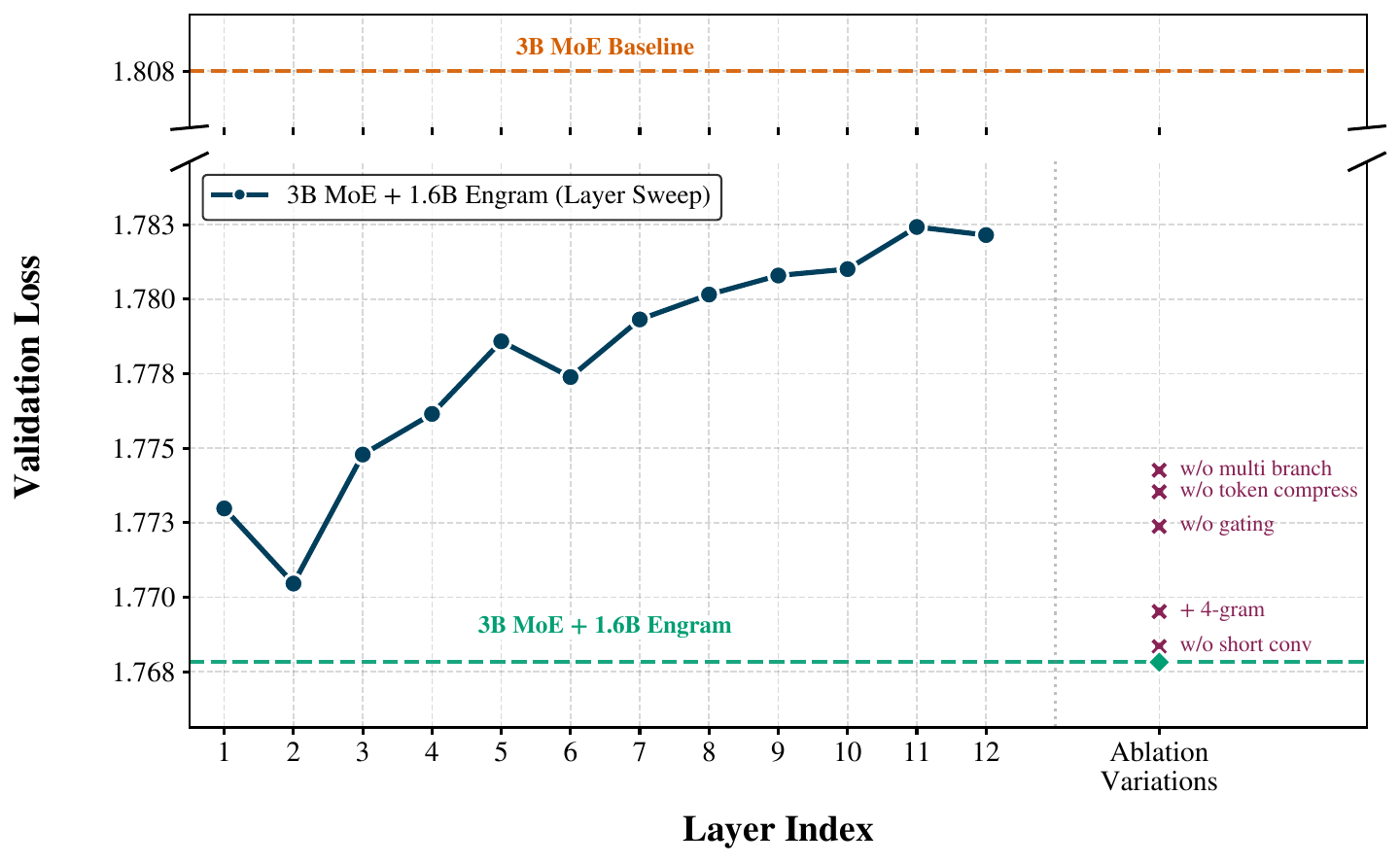}
\caption{
    \textbf{Architecture ablation results.}
    We compare the 3B MoE baseline against Engram variations in two settings:
    (1) Layer Sensitivity (dark blue curve): Sweeping the insertion depth of a single Engram module confirms that early injection (Layer 2) is optimal, whereas efficacy degrades in deeper layers.
    (2) Component Ablation (Right Markers): Removing sub-modules from the reference configuration demonstrates the importance of multi-branch integration, tokenizer compression, and context-aware gating.
}
\label{fig:arch_ablation}
\end{figure}

\subsection{Structural Ablation and Layer Sensitivity}
\label{sec:ablation_structural}

In this section, we ablate Engram under a controlled setting to investigate the effectiveness of each key module design.
Unless otherwise specified, the backbone is a 12-layer 3B MoE model (0.56B activated parameters) trained for 100B tokens.
\autoref{fig:arch_ablation} reports validation loss.
The dashed orange line denotes the 3B MoE baseline (Val Loss $=1.808$).

\paragraph{Reference configuration.}
We augment the backbone with a fixed 1.6B-parameter Engram memory.
Our reference model uses $\{2,3\}$-grams and inserts Engram at Layers 2 and 6, achieving Val Loss $=1.768$,
a substantial improvement over the MoE baseline ($\Delta=0.04$).
All structural ablations below are defined relative to this reference.

\paragraph{Where should memory be injected?}
To study depth sensitivity, we keep the Engram budget fixed (1.6B) but consolidate it into a single Engram module,
and sweep its insertion layer from 1 to 12 (dark blue ``Layer Sweep'' curve in \autoref{fig:arch_ablation}).
This experiment exposes an inherent trade-off in Engram placement.

\textbf{A placement trade-off.}
Injecting Engram early allows it to offload local pattern reconstruction before the backbone expends computational depth, aligning with the backbone's natural hierarchical processing~\citep{tenney2019bert,ghandeharioun2024patchscopes,li2025echoesbertmodernlanguage,DBLP:conf/coling/JinYHZWH0MMDYDZ25}.
However, this incurs a cost in gating precision: early hidden states have not yet aggregated sufficient global context via attention, and the parallel branches lack the representational divergence required for fine-grained modulation~\citep{xie2025mhcmanifoldconstrainedhyperconnections,zhu2025hyperconnections}.
Consequently, optimal placement requires balancing (i) offloading static local patterns early and (ii) utilizing stronger contextual queries for gating later.

The sweep shows that Layer 2 achieves the best single-layer performance (Val Loss $= 1.770$), outperforming Layer 1 and
degrading as the insertion point moves deeper.
This indicates that one round of attention is already sufficient to provide a meaningfully contextualized $\mathbf{h}_t$ for gating,
while still being early enough to replace the backbone's bottom-layer local aggregation.

While Layer 2 is optimal under a single injection constraint, we find that dividing the same 1.6B memory into two smaller
modules (achieved by reducing the embedding dimension $d_{\text{mem}}$) and placing them at Layers 2 and 6 performs even better (Val Loss $=1.768$).
This layered design reconciles the trade-off by combining early intervention with rich, late-stage contextual gating.
More importantly, layered insertion also provides a practical system advantage, enabling better utilization of the memory hierarchy as discussed in \autoref{sec:system_efficiency}.

\paragraph{Which components matter?}
Starting from the reference configuration, we ablate individual design choices while
keeping the Engram parameter budget fixed. Results are denoted by markers in \autoref{fig:arch_ablation}.
We find that three components yield the most significant gains:
(i) branch-specific fusion within the multi-branch backbone,
(ii) context-aware gating, and
(iii) tokenizer compression.
Removing any of these causes the largest regressions in validation loss.
Specifically, for the ``w/o multi branch'' ablation, we retain the mHC backbone structure but replace the branch-specific gating with a single Engram fusion applied to the hidden states after the pre-mapping $\mathcal{H}^{pre}$~\citep{xie2025mhcmanifoldconstrainedhyperconnections}.

Other changes have smaller effects: removing the lightweight depthwise convolution only marginally degrades performance.
Allocating capacity to 4-grams is slightly suboptimal under a fixed 1.6B budget—likely because it dilutes capacity
from the more frequent 2/3-gram patterns—though we do not rule out that higher-order $N$-grams become beneficial at larger memory scales.

\subsection{Sensitivity Analysis}
\label{sec:dse_sensitive}

\begin{figure}[t]
\centering
\includegraphics[width=0.9\linewidth]{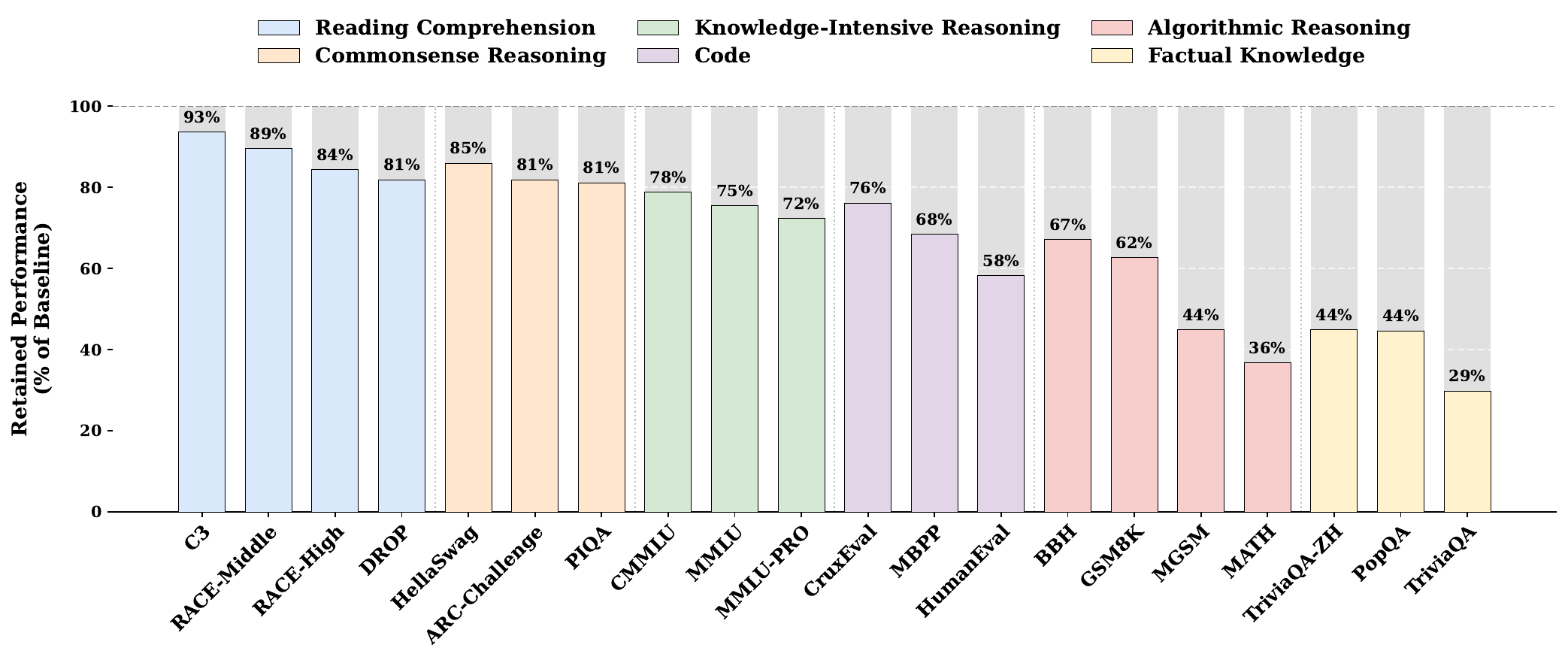}
\caption{\textbf{Retained performance under Engram ablation.} Factual knowledge relies heavily on the Engram module, whereas reading comprehension is largely preserved by the backbone.}
\label{fig:sensitivity_analysis}
\end{figure}

To characterize the functional contribution of the Engram module, we evaluate the model by completely suppressing the sparse embedding output during inference while keeping the backbone unchanged. Crucially, this post-hoc ablation induces a \textbf{training--inference inconsistency}, potentially introducing noise in complex, mixed-capability tasks. Consequently, we prioritize the analysis of \textit{Factual Knowledge} and \textit{Reading Comprehension}—the two extremes of the sensitivity spectrum—which exhibit the highest signal-to-noise ratio under this stress test. 

As shown in \autoref{fig:sensitivity_analysis}, the results reveal a sharp functional dichotomy. Factual knowledge benchmarks suffer a catastrophic collapse, retaining only 29--44\% of the original performance (e.g., TriviaQA at 29\%), confirming that the Engram module acts as the primary repository for parametric knowledge. Conversely, reading comprehension tasks are remarkably resilient, retaining 81--93\% (e.g., C3 at 93\%), suggesting that context-grounded tasks rely primarily on the backbone's attention mechanism rather than Engram.

\subsection{System Efficiency}
\label{sec:inference_analysis}

A pivotal system advantage of Engram over routing-based MoE is that its sparse activations are addressed by explicit, static hash IDs. This yields a strictly deterministic memory access pattern: indices for the next Engram lookup are fixed once the token sequence is known and can be computed before the corresponding layer executes. 

\paragraph{Experimental Setup.}
We implemented an inference harness based on nano-vLLM\footnote{\url{https://github.com/GeeeekExplorer/nano-vllm}}—a streamlined prototype of the industry-standard vLLM engine~\citep{kwon2023efficient}. To obtain a clean latency baseline without the confounding communication patterns of Expert Parallel in MoE, we benchmark on two dense backbones (Dense-4B and Dense-8B).
We insert a massive 100B-parameter Engram layer into the second Transformer block, with the entire embedding table resident in host DRAM.
During inference, the system prefetches embeddings for the Engram layer asynchronously, overlapping the PCIe transfer with the computation of the first block.

\begin{table}[htbp]
    \centering
    \caption{\textbf{End-to-end Inference Throughput}. We measure infernece throughput with a 100B-parameter Engram layer entirely offloaded to host memory.}
    \label{tab:vllm}
    \vspace{2mm}
    \begin{tabular}{llc}
        \toprule
        \multicolumn{3}{c}{\textbf{Experimental Setup}} \\
        \midrule
        \multicolumn{2}{l}{\textit{Hardware}} & NVIDIA H800 \\
        \multicolumn{2}{l}{\textit{Workload}} & 512 Sequences \\
        \multicolumn{2}{l}{\textit{Sequence Length}} & Uniform$(100, 1024)$ \\
        \midrule
        \multicolumn{3}{c}{\textbf{Throughput Results}} \\
        \midrule
        \textbf{Base Model} & \textbf{Configuration} & \textbf{Throughput (tok/s)} \\
        \midrule
        \multirow{2}{*}{4B-Dense} 
            & Baseline & 9,031.62 \\
            & + 100B Engram (CPU Offload) & 8,858.28 \\
        \midrule
        \multirow{2}{*}{8B-Dense} 
            & Baseline & 6,315.52 \\
            & + 100B Engram (CPU Offload) & 6,140.02 \\
        \bottomrule
    \end{tabular}
\end{table}

\paragraph{Results.}
As detailed in \autoref{tab:vllm}, offloading a 100B-parameter embedding table incurs a negligible throughput penalty, peaking at only $2.8\%$ on the 8B backbone.
This confirms that the compute intensity of early dense blocks provides a sufficient temporal window to mask the retrieval latency. Crucially, the effective communication volume per step scales with the number of activated slots rather than the total embedding table size.

Crucially, this experiment serves as a conservative baseline. While the hierarchical design in \autoref{sec:system_efficiency} exploits Zipfian locality to cache frequent items in HBM, our experimental setup forces \textit{all} retrievals to traverse the PCIe bus from host memory. The fact that this baseline retrieval strategy yields minimal overhead strongly suggests that a fully optimized, locality-aware implementation would incur negligible throughput penalty.

\subsection{Case Study: Gating Visualization}
\label{sec:case_study}

\begin{figure}[t]
    \centering
    \includegraphics[width=1.0\linewidth]{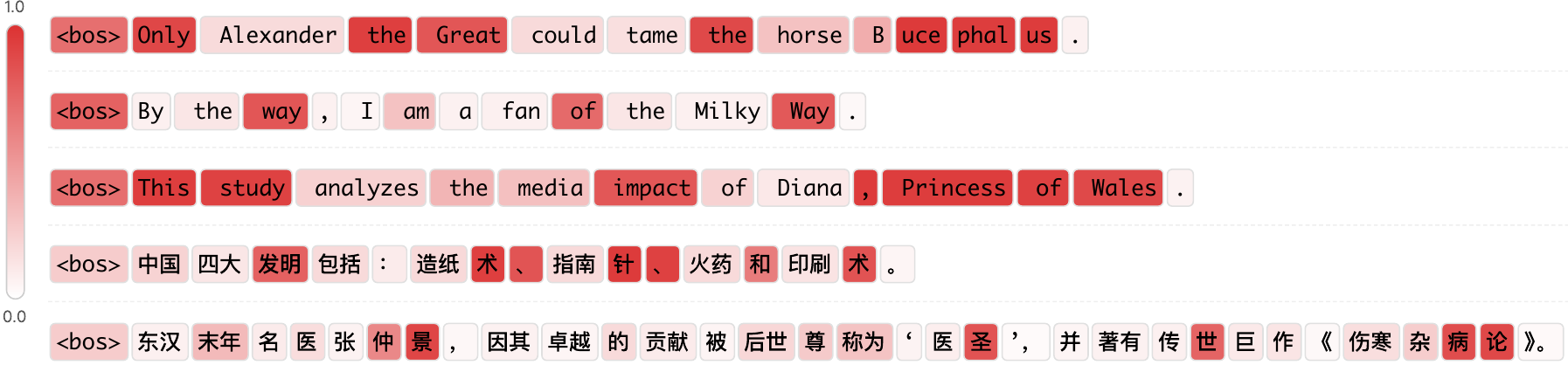}
    \caption{
        \textbf{Visualization of the gating mechanism of Engram.} The heatmap intensity corresponds to the magnitude of the gating scalar $\alpha_t \in [0, 1]$, where darker red indicates stronger activation. Because Engram operates on suffix $N$-grams (here $N=3$), a high activation on a specific token $x_t$ implies that the preceding tokens culminating in that token (e.g., the phrase ending at $t$) are recognized as a static pattern effectively retrieved from memory.
    }
    \label{fig:visualization}
\end{figure}

In \autoref{sec:engram_fusion}, we introduced the context-aware gating mechanism, designed to dynamically modulate the integration of retrieved static memory into the backbone. To empirically validate whether Engram behaves as intended, we visualize the gating scalar $\alpha_t$ of Engram-27B\footnote{As detailed in our architecture setup, this model utilizes a mHC ($M=4$) with Engram modules inserted at layers 2 and 15. Consequently, for any given token, the model computes a total of 8 distinct gating scalars. We observe that not every branch encodes interpretable activation patterns. For the clarity of this visualization, we select and display the gating values most strongly correlated with semantic pattern matching.} across various samples in \autoref{fig:visualization}.

The results demonstrate a distinct pattern of selectivity. The gating mechanism consistently activates (shown in red) upon completing local, static patterns. In English, we observe strong activations on multi-token named entities (e.g., ``Alexander the Great'', ``the Milky Way'') and formulaic phrases (e.g., ``By the way'', ``Princess of Wales''). This behavior generalizes effectively across languages. In the Chinese examples, Engram identifies and retrieves distinct idiomatic expressions and historical entities, such as ``Four Great Inventions'' (\textbf{四大发明}) and ``Zhang Zhongjing'' (\textbf{张仲景}). These qualitative results confirm that Engram successfully identifies and handles stereotyped linguistic dependencies, effectively relieving the Transformer backbone from memorizing these static associations.

%% file: sections/related_work.tex
\section{Related Work}

\paragraph{$N$-gram Modeling and Embedding Scaling.}
Originating from Shannon's framework~\citep{shannon1948mathematical}, $N$-gram models rely on local history to predict tokens, traditionally employing smoothing techniques~\citep{kneser1995improved,DBLP:journals/tsp/Katz87} to mitigate data sparsity. Despite the paradigm shift toward neural architectures~\citep{bengio2003neural} for capturing long-range dependencies, the computational efficiency of $N$-gram lookups has been preserved in modern representation learning, as exemplified by seminal works like FastText~\citep{bojanowski2017enriching}.

Recently, this paradigm has resurged as \textit{embedding scaling}. While architectures such as Per-Layer Embeddings~\citep{gemma_3n_2025}, STEM~\citep{sadhukhan2026stem}, L3~\citep{tseng20263} and DeepEmbed~\citep{rwkv_deepembed_wiki_2025} expand capacity via massive tables, a distinct line of pioneering research—most relevant to our approach—integrates compositional $N$-gram structures directly into the representation space. 
N-Grammer~\citep{roy2022n} directly augments the transformer architecture with $N$-gram that are constructed from a discrete latent text representation, while \citet{feng2023memory} demonstrates the effectiveness of $N$-gram for speech recognition.
SuperBPE~\citep{liu2025superbpe} and SCONE~\citep{yu2025scaling} explicitly target high-frequency patterns: the former by merging multi-word expressions into ``superword'' tokens, and the latter via an auxiliary encoding model. 
In parallel, OverEncoding~\citep{huang2025over} and Byte Latent Transformer (BLT)~\citep{pagnoni2025byte} adopt hash $N$-gram embeddings to capture local dependencies at the token and byte levels, respectively. These studies collectively demonstrate the efficacy of scaling parameters through $N$-gram representations with minimal computational overhead. While these approaches offer significant gains in their respective settings, our work diverges fundamentally in two key dimensions. 
\begin{itemize}
    \item First, regarding modeling and evaluation protocols. Prior approaches often treat $N$-gram embeddings as external augmentations without validating their efficiency under strictly fair comparison protocols. For instance, SCONE \citep{yu2025scaling} is inference-focused and relies on auxiliary modules that incur additional training FLOPs. Similarly, OverEncoding \citep{huang2025over} fails to yield meaningful improvements on sparse MoE backbones even under a non-isoparametric setting. In contrast, we treat conditional memory as a first-class modeling primitive instantiated via the carefully designed Engram module. By rigorously evaluating this design within our \textit{Sparsity Allocation} framework, we demonstrate its clear advantage over strictly iso-parameter and iso-FLOPs MoE baselines.
    \item  Second, from a system perspective, we advocate for algorithm-system co-design. Existing approaches place embeddings strictly at the input layer (Layer 0), which inherently serializes memory access and computation~\citep{yu2025scaling,huang2025over}. Engram, conversely, strategically injects memory into deeper layers to enable communication-computation overlap. Furthermore, by exploiting the inherent Zipfian distribution of $N$-grams, we could maximize the utility of the hardware memory hierarchy. This holistic design allows Engram to scale to massive parameters with negligible inference overhead.
\end{itemize}

\paragraph{High-Cardinality Categorical Embeddings.}
A closely related representation problem arises in large-scale recommender systems, where models learn embeddings for hundreds of categorical features whose vocabularies may contain millions to billions of IDs~\citep{coleman2023unified}.
Prior work improves the parameter--accuracy and memory-access trade-offs through multi-hash and compositional representations~\citep{tito2017hash,shi2020compositional},
frequency-aware hashing and collision management
~\citep{zhang2020model,tsang2022clustering,zhang2024cafe},
frequency- or importance-adaptive capacity allocation
~\citep{ginart2021mixed,joglekar2020neural,liu2021learnable},
and structured compression based on quantization, tensor decomposition,
or shared memory layouts
~\citep{kang2020learning,yin2021tt,desai2021random}.
Other approaches share representations across multiple categorical fields
~\citep{coleman2023unified}, generate embeddings without explicit tables
~\citep{kang2021learning}, or replace arbitrary item IDs with learned
semantic identifiers
~\citep{rajput2023recommender,singh2024better}.
Engram shares with this literature the challenge of representing an
extremely large discrete key space under a highly skewed access
distribution, but differs in constructing keys from ordered textual
$N$-grams and injecting the retrieved representations into intermediate
Transformer layers through context-aware gating.

\paragraph{Mixture-of-Experts.} MoE architectures decouple model capacity from computational cost by conditionally activating a sparse subset of experts per token, a paradigm introduced by~\citet{shazeer2017outrageously}. Subsequent innovations such as GShard~\citep{lepikhin2020gshard}, BASE~\citep{pmlr-v139-lewis21a}, Switch Transformer~\citep{fedus2022switch} and GLaM~\citep{du2022glam} enabled super-linear parameter scaling while maintaining constant inference costs. More recently, DeepSeek-MoE~\citep{dai2024deepseekmoe} demonstrated superior efficiency, significantly outperforming dense models with equivalent active parameters via fine-grained expert segmentation and shared expert isolation. Adopting this architecture, state-of-the-art models such as DeepSeek-V3~\citep{liu2024deepseek} and Kimi-k2~\citep{kimi-k2} have further pushed total parameters to hundreds of billions scale.
 
\paragraph{Memory Network.}
Research on memory-augmented networks aims to expand model capacity without a proportional increase in computational cost, broadly categorized into parametric and non-parametric approaches. Parametric memory methods, such as PKM~\citep{lample2019large}, PEER~\citep{he2024mixture}, Selfmem~\citep{cheng2023lift}, Memory+~\citep{berges2024memory} and UltraMem~\citep{huang2024ultra,huang2025ultramemv2}, integrate large-scale, sparse key-value stores directly into the model layers, thereby significantly increasing capacity with negligible impact on FLOPs. 
Conversely, non-parametric memory approaches like REALM~\citep{guu2020retrieval}, RETRO~\citep{borgeaud2022improving,wang2023shall}, CoG~\citep{lan2023copy,caoretrieval} and PlugLM~\citep{cheng2023decouple} decouple knowledge storage from model processing, treating the external memory as an editable and scalable key-value store that allows the model to adapt to evolving information without expensive retraining.

\paragraph{Mechanisms of Knowledge Storage.}
Parallel to capacity scaling, substantial research has scrutinized the internal mechanisms governing how Transformers encode and retrieve factual knowledge. The Feed-Forward Networks (FFNs) are widely hypothesized to function as Key-Value memories~\citep{geva2021transformer}. Under this framework, the first layer acts as a pattern detector ("keys") while the second layer projects specific information into the residual stream ("values"). This modularity is evidenced by the identification of specific ``knowledge neurons'' responsible for storing distinct facts~\citep{dai2022knowledge}. Further validation is provided by causal tracing methodologies, which map the information flow of factual recall to specific FFN layers~\citep{meng2022locating}. These insights have enabled precise model editing algorithms such as ROME~\citep{meng2022locating} and MEMIT~\citep{meng2022mass}, which allow for the direct update of factual associations without retraining. Moreover, investigations into internal representations, such as those in Othello-GPT~\citep{li2024emergentworldrepresentationsexploring}, suggest that these storage mechanisms may facilitate the emergence of structured ``world models'' rather than mere statistical memorization.

%% file: tabs/appendix_detailed_arch.tex
\begin{table}[htbp]
\centering
\resizebox{0.85\linewidth}{!}{%
\begin{tabular}{@{}l cccc@{}}
\toprule
 & \textbf{Dense-4B} & \textbf{MoE-27B} & \textbf{Engram-27B} & \textbf{Engram-40B} \\
\midrule
Total Params & 4.1B & 26.7B & 26.7B & 39.5B \\
Active Params & \multicolumn{4}{c}{3.8B} \\
Total Tokens & \multicolumn{4}{c}{262B} \\
\midrule
Layers & \multicolumn{4}{c}{30} \\
Dimension & \multicolumn{4}{c}{2560} \\
Leading Dense Layers & - & 1 &1 &1 \\
Routed Experts & - & 72& 55 & 55 \\
Active Experts & - & 6 & 6 & 6 \\
Shared Experts & - &2 & 2& 2\\
Load Balancing Method & - & \multicolumn{3}{c}{Loss Free~\citep{wang2024auxiliarylossfreeloadbalancingstrategy}} \\
\midrule
Attention module & \multicolumn{4}{c}{MLA~\citep{deepseekai2024deepseekv2strongeconomicalefficient}} \\ 
RoPE $\theta$ & \multicolumn{4}{c}{10000} \\
mHC Expansion Rate & \multicolumn{4}{c}{4} \\
Sequence Length & \multicolumn{4}{c}{4096} \\
Vocab Size & \multicolumn{4}{c}{129280} \\
Batch Size & \multicolumn{4}{c}{1280} \\
Training Steps & \multicolumn{4}{c}{50000} \\
Backbone Optimizer & \multicolumn{4}{c}{Muon~\citep{jordan2024muon}} \\
Embedding Optimizer & \multicolumn{4}{c}{Adam~\citep{kingma2014adam}} \\
Base Learning Rate & \multicolumn{4}{c}{4e-4} \\
Lr Scheduler & \multicolumn{4}{c}{Step Decay~\citep{bi2024deepseek}} \\
Weight Decay & \multicolumn{4}{c}{0.1}  \\
\midrule
Engram Dim $d_{\text{mem}}$ & - & - & 1280 & 1280 \\
Engram Vocab Size & - & - & 2262400 & 7239680 \\
Engram Num Head & - & - & 8 & 8 \\
Engram Layer & - & - & [2,15] & [2,15] \\
Engram $N$-gram & - & - & [2,3] & [2,3] \\
Engram combine mHC & - & - & True & True \\
Engram tokenizer compression  & - & - & True & True \\
Engram Conv Zero Init  & - & - & True & True \\
Engram Lr Multipler  & - & - & x5 & x5 \\
Engram Weight Decay  & - & - & 0.0 & 0.0 \\
Engram Optimizer (Embed. only)  & - & - & \multicolumn{2}{c}{Adam~\citep{kingma2014adam}}  \\
\bottomrule
\end{tabular}
}
\caption{Detailed model architecture information and training hyper parameters.}
\label{tab:detailed_arch}
\end{table}